\PassOptionsToPackage{table}{xcolor}
\documentclass{article}

\usepackage[preprint]{neurips_2025}

\usepackage[utf8]{inputenc} 
\usepackage[T1]{fontenc}    
\usepackage{amsfonts}       
\usepackage{nicefrac}       
\usepackage{microtype}      

\usepackage{graphicx}       
\usepackage{microtype}      
\usepackage{hyperref}       
\usepackage{url}            
\usepackage{amsmath}        
\usepackage{amssymb}        
\usepackage{enumitem}       
\usepackage{xspace}         

\usepackage{booktabs}       
\usepackage{array}          
\usepackage{tabularx}       
\usepackage{longtable}      
\usepackage{siunitx}        
\usepackage{subfig}         

\usepackage{tcolorbox}      
\usepackage{xcolor}  
\usepackage{multirow}       
\usepackage{wrapfig}        
\usepackage{lineno}         
\usepackage{lipsum}         
\usepackage{nicematrix}     

\hypersetup{colorlinks,citecolor={blue}}

\usepackage{listings}
\usepackage{listings-ext}
\usepackage{textcomp}
\newcommand{\threebackticks}{\textasciigrave\textasciigrave\textasciigrave}

\newcommand{\ourdataset}{\textsc{OpenCodeReasoning-II}\xspace}

\newcommand{\bplus}[1]{{\footnotesize\textcolor{blue}{(+\text{#1})}}}

\title{\ourdataset: A Simple Test Time Scaling Approach via Self-Critique}

\author{Wasi Uddin Ahmad, Somshubra Majumdar, Aleksander Ficek, Sean Narenthiran, \\ [2pt]
{\bf Mehrzad Samadi, Jocelyn Huang, Siddhartha Jain, Vahid Noroozi, Boris Ginsburg} \\ [2pt]
NVIDIA \\ [2pt]
Santa Clara, CA 95051, USA \\ [2pt]
\texttt{\{wasiuddina, smajumdar, aficek, snarenthiran\}@nvidia.com} \\ [2pt]
\texttt{\url{https://huggingface.co/datasets/nvidia/OpenCodeReasoning-2}} 
}

\begin{document}

\setlength{\abovedisplayskip}{5pt}
\setlength{\belowdisplayskip}{5pt}

\maketitle

\begin{abstract}
Recent advancements in reasoning-based Large Language Models (LLMs), particularly their potential through test-time scaling, have created significant opportunities for distillation in code generation and critique. However, progress in both areas fundamentally depends on large-scale, high-quality datasets. In this work, we introduce \ourdataset, a dataset consists of 2.5M question-solution-critique triples ($\approx$ 35K unique programming questions), making it nearly twice the size of the previous largest publicly available code reasoning dataset. In this work, we employ a two-stage supervised fine-tuning strategy. The first stage focuses on fine-tuning for code generation, while the second stage involves the joint training of models for both code generation and critique. Our resulting finetuned \texttt{Qwen2.5-Instruct} models achieve performance in code generation that either exceeds or equals the best prior open-weight distilled models. Notably, the integration of our code generation and critique models leads to significant improvements in competitive coding performance. Furthermore, we present an extension of the LiveCodeBench benchmark to specifically support the C++ programming language, thereby facilitating more comprehensive LLM evaluation using this benchmark.
\end{abstract}

\section{Introduction}
\label{sec:introduction}

Large language models (LLMs) have undergone rapid advancements in recent years, from chain-of-thought (CoT) prompting \citep{wei2022chain}, followed by System-2 reasoning that employs test-time compute scaling approaches \citep{li202512surveyreasoning, zhang2025surveytesttimescalinglarge}. Test-time scaling has allowed LLMs to dedicate more computational resources during inference to perform logical reasoning. The release of models like DeepSeek-R1 \citep{deepseek_r1}, which demonstrated impressive reasoning capabilities, has spurred increased interest in distilling these test-time compute capabilities into smaller fine-tuned models. This pursuit is driven by the observation that providing LLMs with more computational power during inference directly translates to tangible improvements in their outputs, particularly on complex tasks like competitive coding.

As models have improved on reasoning tasks by leveraging test-time compute, a natural research question emerges: \textit{how can test-time compute be effectively scaled?} Inference time scaling has  been found to emerge from applying reinforcement learning on domains with verifiable outcomes such as with math and coding \citep{setlur2025scalingtesttimecomputeverification, qu2025optimizingtesttimecomputemeta, yu2025dapoopensourcellmreinforcement}. Other works have found benefits by scaling inference compute by using repeated sampling in parallel \citep{wang2023selfconsistencyimproveschainthought, brown2024largelanguagemonkeysscaling, wu2025inferencescalinglawsempirical}. After sampling multiples times the best solution can be selected using a variety of methods such as majority voting, reward models or LLM-as-a-judge \citep{chen2024llmcallsneedscaling, liu2025acemathadvancingfrontiermath, zeng2025acecoderacingcoderrl, moshkov2025aimo2winningsolutionbuilding}. Recent works have found that scaling test-time compute also results in better critique models in the form of reasoning-based judges or Generative Reward Models (GenRM's) \citep{mahan2024generativerewardmodels, zhang2025generativeverifiersrewardmodeling, liu2025inferencetimescalinggeneralistreward}. Additionally, \citet{wang2025critiquefinetuninglearningcritique} introduces Critique Fine-Tuning (CFT), a method that uses model-based critiques to more effectively distill reasoning capabilities than standard Supervised-Finetuning (SFT) for math problems. Our work unifies Critique Fine-Tuning (CFT) with reasoning data distillation to effectively scale test-time compute and enhance coding capabilities.

Reasoning-based data distillation has proven to be a powerful technique for enhancing coding performance, often without requiring reinforcement learning \citep{bespoke_stratos, penedo2025codeforces, openthoughts, li2025llmseasilylearnreason}. Follow-up works have found continual gains from supervised fine-tuning on increasingly larger CoT reasoning datasets \citep{xu2025kodcodediversechallengingverifiable, ahmad2025opencodereasoningadvancingdatadistillation}. Despite the recognized benefits of critique fine-tuning and applying test-time compute to generative reward models, a significant gap persists: there are no publicly available coding datasets that specifically include reasoning-based CoT critiques. Existing datasets with critique data feature non-reasoning critiques, execution test pass-rates, or focus on different domains \citep{ahmad2025opencodeinstructlargescaleinstructiontuning, ahmad2025opencodereasoningadvancingdatadistillation, zeng2025acecoderacingcoderrl, zhang2025generativeverifiersrewardmodeling, wang-etal-2024-helpsteer}.


\begin{figure*}[t]
\centering
\includegraphics[width=\textwidth]{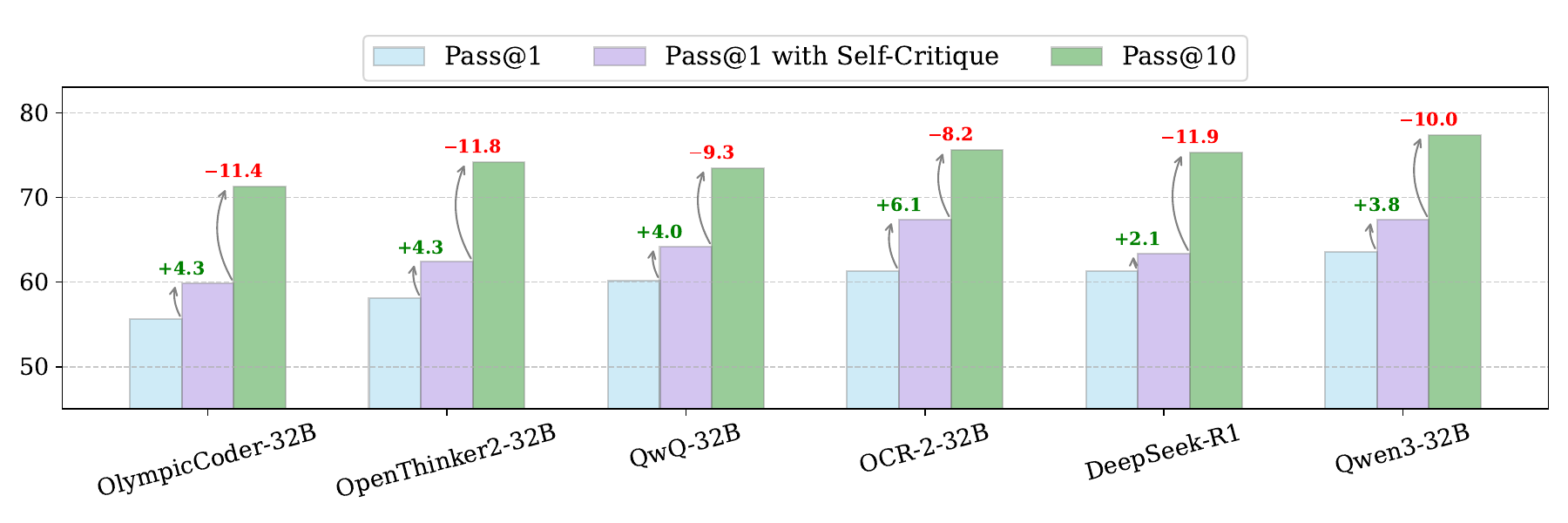}
\caption{Demonstrating performance gains on LiveCodeBench, achieved through test-time scaling by generating 10 solutions per problem and employing self-critique for selecting the final output. Self-critique led to the greatest performance boost in our finetuned model, \textbf{OCR-2-32B}.}
\label{fig:intro_plot}
\end{figure*}

To bridge this gap and further leverage test-time scaling, we present \ourdataset, the largest publicly accessible code reasoning dataset created to date. This dataset comprises 2.5 million question-solution-critique triples originating from roughly 35,000 distinct programming problems. Both solutions and critiques are structured as reasoning CoTs, offering detailed justifications for solution generation and validation. Using this dataset, we trained models with 7B, 14B, and 32B parameters using a two-stage finetuning approach. We conduct evaluation under parallel test-time scaling and demonstrated significant improvements compared to open-weight models, as illustrated in \autoref{fig:intro_plot}. Furthermore, we contribute an extension to LiveCodeBench, with a specific focus on supporting LLM evaluation using C++. The contributions of this work can be summarized as follows:


\begin{enumerate}[leftmargin=*]
\item We introduce \ourdataset, a large-scale dataset containing 1.4 million Python and 1.1 million C++ solutions, along with their corresponding critique labels, detailed reasoning traces, and execution pass rates, all derived from 35,000 unique programming questions.

\item With the aim of facilitating more comprehensive LLM evaluation in the C++ programming language, we have extended LiveCodeBench to incorporate C++ support. This enhanced benchmark is now publicly available to promote further research and development.

\item We show the effectiveness of \ourdataset by fine-tuning \texttt{Qwen2.5-Instruct} models in a two-stage process. This fine-tuning enabled our models to achieve code generation performance that surpasses or matches the leading prior open-weight distilled models. Furthermore, integrating code generation with critique through a simple test-time scaling strategy resulted in substantial gains on the LiveCodeBench benchmark, in both Python and C++.

\item We perform an in-depth analysis to provide insights on the opportunities in self-critique methods under test-time scaling, impact of data scaling, transfer between Python and C++ languages. 
\end{enumerate}







\section{Development of \ourdataset and LiveCodeBench-C++}
\label{sec:data}

\begin{figure*}[t]
\centering
\includegraphics[width=\textwidth]{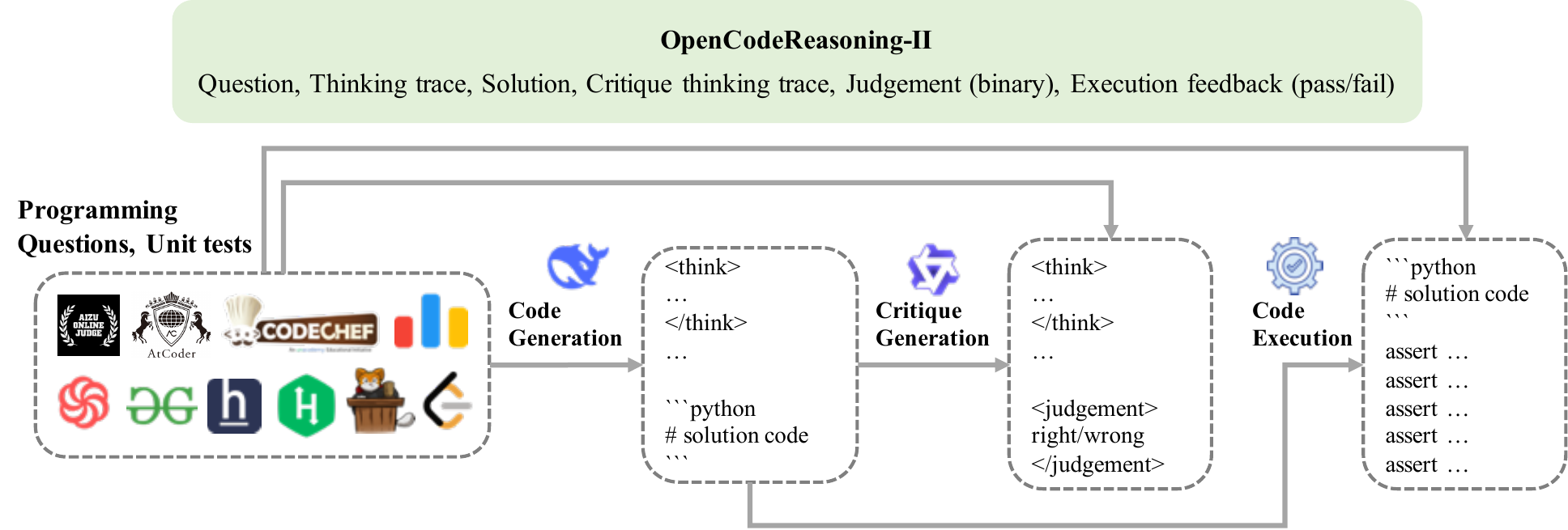}
\caption{Overview of the \ourdataset development stages.}
\label{fig:ocr2_overview}
\end{figure*}

\subsection{Construction of \ourdataset}
The construction of the \ourdataset dataset involved a four-stage approach which is demonstrated in \autoref{fig:ocr2_overview}. The initial stage consisted of compiling a diverse collection of competitive coding problems with unit tests from various origins. Following this, a large language model (LLM) equipped with reasoning capabilities was leveraged to produce corresponding solutions. Next, a reasoning-capable LLM was employed to generate critique for these solutions. In the final stage, we obtained execution results for a portion of the generated solutions. During stage-II and stage-III, the generated solutions and their corresponding critiques underwent a post-processing stage aimed at guaranteeing their structural consistency.

\subsubsection{Programming Questions Collection}

To create our dataset, we drew problems from the TACO corpus \citep{li2023taco}, the APPS benchmark \citep{apps2021}, the CodeContests collection \citep{li2022codecontests}, and CodeForces problems through the OpenR1 initiative \citep{openr1_codeforces}. Due to the frequent overlap found in public datasets, we applied a fuzzy matching-based de-duplication method, leading to a final set of 34,799 unique questions of diverse difficulty. The distribution of these questions is detailed in \autoref{tab:data_stat}.

\paragraph{Contamination Assessment}
To ensure the integrity of \ourdataset, we rigorously investigated potential data leakage between our collected programming questions and major code generation evaluation suites \citep{jain2025livecodebench, li2022codecontests, chen2021codex_humaneval, austin2021mbpp}. Our methodology mirrored the protocol outlined by \cite{yang2023lmsys_decontaminate}, which involved computing the cosine similarity (with a cutoff of 0.7) to identify the closest counterpart within the benchmark datasets for every distinct question in \ourdataset. We used Llama-3.3-70B-Instruct \citep{grattafiori2024llama} as a judge to assess semantic similarity, and it identified 674 questions that potentially overlap with evaluation benchmarks. Following this decontamination, we proceeded to generate solutions for the remaining 34,125 programming questions.

\subsubsection{Solution Generation using DeepSeek-R1}
In this stage, we generated multiple solutions for each question leveraging the DeepSeek-R1 model \citep{deepseek_r1}. These solutions were generated in Python and C++ and sampled using Nucleus Sampling \citep{holtzman2020nucleus_sampling} with temperature $0.6$ and top-p $0.95$. We utilized SGLang \citep{zheng2024sglang} for this generation process, allowing for a maximum output sequence length of 32k tokens. The prompt used to generate solutions is provided in \autoref{fig:combined_data_gen}.

\paragraph{Post-processing and Filtering}
We post-processed and filtered the generated responses to ensure the required information were present in the output. Initially, we checked if each responses contained reasoning traces enclosed by the \textit{\textless think\textgreater} and \textit{\textless /think\textgreater} tags. Next, we extracted the solution segments, separating the reasoning traces from the rest of the response. We then confirmed the presence of code blocks delimited by \threebackticks{}python~\ldots\threebackticks or \threebackticks{}cpp~\ldots\threebackticks. Finally, we used Tree Sitter \citep{tree_sitter} to validate the syntactic correctness of these code blocks. Notably, these filtering procedures led to the removal of a very few responses.

\begin{table}[t]
\centering
\def\arraystretch{1.1}%
\begin{tabular}{l |r r|r r}
\toprule
\multirow{2}{*}{Source} & \multicolumn{2}{c|}{Python} & \multicolumn{2}{c}{C++} \\ 
& \# Question & \# Sample & \# Question & \# Sample \\ 
\midrule
AIZU            & 2151      & 71,681    & 2067      & 35,471    \\ 
AtCoder         & 2080      & 64,468    & 1988      & 62,493    \\ 
CodeChef        & 3869      & 120,040   & 3830      & 171,882   \\ 
CodeForces      & 15641     & 834,523   & 11887     & 355,180   \\ 
Codewars        & 2506      & 79,771    & 2492      & 155,162   \\ 
GeeksForGeeks   & 2670      & 58,154    & 2668      & 167,610   \\ 
HackerEarth     & 2285      & 73,559    & 2273      & 82,765    \\ 
HackerRank      & 912       & 26,106    & 903       & 43,867    \\ 
Kattis          & 1235      & 39,938    & 1209      & 49,699    \\ 
LeetCode        & 777       & 29,926    & 775       & 50,346    \\ 
\midrule
Total           & 34,125    & 1,398,166 & 30,092    & 1,174,475 \\ 
\bottomrule
\end{tabular}
\vspace{2mm}
\caption{Number of questions and corresponding samples in \ourdataset, spanning across ten programming platforms.}
\vspace{-4mm}
\label{tab:data_stat}
\end{table}


\subsubsection{Critique Generation using QwQ-32B}

In this stage, we prompted QwQ-32B \citep{qwq32b} to to generate critiques for the programming questions and their corresponding code solutions. The specific prompt used to generate critique is detailed in \autoref{fig:critique_data_gen}. We employed the same generation settings as for solution generation: temperature-based nucleus sampling via SGLang with a maximum output length of 24k tokens. Following a similar post-processing and filtering approach to solution generation, we verified that each critique contained reasoning traces (within \textit{\textless think\textgreater} and \textit{\textless /think\textgreater} tags) and a final judgment (within \textit{\textless judgment\textgreater} and \textit{\textless /judgment\textgreater} tags). We retain responses only if their final judgment is binary: either \texttt{right} or \texttt{wrong}. Otherwise, we discard them. The rationale behind this choice stems from our preliminary experiments. When evaluating reasoning-enabled LLMs (R1 and QwQ-32B) with binary, categorical (correct, partially correct, incorrect), and numeric (1-5) judgment options, we observed a significant tendency for the models to favor binary responses (such as 1 or 5, or correct/incorrect). Consequently, we adopted binary judgment generation for our critique responses.

\subsubsection{Verifying Solutions with Unit Tests}
In the final stage of \ourdataset construction, we executed the generated code solutions against their corresponding unit tests, which were collected alongside the questions from public benchmarks. To ensure meaningful and manageable execution outputs, we selected a subsample of \ourdataset where each question had at least 5  unit tests, with a maximum of 50 randomly selected if more were available. This subsample comprised 60\% of \ourdataset. Following execution, we calculated the \texttt{pass rate}, which is included in the public release of \ourdataset. We expect these execution outputs to facilitate future research, including the application of offline reinforcement learning for LLM improvement.

\subsection{Extending LiveCodeBench for C++}

While LiveCodeBench \citep{jain2025livecodebench} aims to provide a contamination-free evaluation of LLMs for code, its limitation to Python hindered our ability to assess LLMs on C++, a widely used language in competitive coding. 
To address this, we extended LiveCodeBench to include C++. We selected problems from \texttt{release\_v5} within the date range of 2408 to 2502, resulting in 279 problems (175 from AtCoder and 104 from LeetCode). 
Notably, AtCoder problems utilize standard input/output for testing, whereas LeetCode problems provide starter code, requiring function invocation for evaluation. We collected the C++ starter code for LeetCode problems and adapted their test cases to enable evaluation in our extended benchmark.
The dataset is publicly available at \url{https://huggingface.co/datasets/nvidia/LiveCodeBench-CPP}.

\section{A Simple Test-time Scaling Approach via Self-Critique}
\label{sec:method}

To showcase the potential of \ourdataset, we establish a straightforward self-critique-based test-time scaling approach as a simple baseline for future research. This section outlines our fine-tuning methodology and the subsequent inference setup for test-time scaling.

\paragraph{Finetuning Setup}
We fine-tuned the \texttt{Qwen2.5-Instruct} models in two stages. Stage I involved fine-tuning for code generation, followed by Stage II where we jointly fine-tuned for both code generation and self-critique. We used the same prompts for fine-tuning as those employed for data generation (illustrated in \autoref{fig:combined_data_gen} and \autoref{fig:critique_data_gen}). The models underwent three epochs of fine-tuning in Stage I and one epoch in Stage II. While this work presents our initial approach, we plan to investigate more advanced fine-tuning techniques in the future.

\paragraph{Inference Setup for Test-Time Scaling}
At inference time, we prompt the fine-tuned models to first produce a solution to a programming question and then to critique their own output (self-critique). This process facilitates parallel scaling \citep{zeng2025revisitingtesttimescalingo1like}, where multiple solutions are generated concurrently, and the best is chosen as the final result. The success of parallel scaling hinges on two factors: (1) coverage, the probability of generating at least one correct solution, and (2) the selection method's accuracy in identifying a correct solution if one or more solutions are labeled as correct. In this study, we evaluate coverage using \texttt{pass@k} and selection efficacy using \texttt{pass@1|select@k} (which we term as \texttt{critique@k}). A limitation of our work is the binary nature of the self-critique judgments, which necessitates a strategy for selecting the best solution among multiple (\texttt{right}) generations. Motivated by the findings of \cite{wang2025thoughtsunderthinking}, which found that the longer reasoning traces often correlate with incorrect final solutions, we utilize a simple heuristic to choose the \texttt{right} solution from a pool of "right"-labeled candidates: select the solution with the \textit{shortest} critique reasoning trace. A detailed comparison against a randomized selection approach is provided in \autoref{sec:critique_selection_ablation}. Recognizing the naivety of this method, we leave the investigation of more sophisticated selection techniques for future work.

\section{Main Evaluation}
\label{sec:results}

\begin{table}[t]
\centering
\def\arraystretch{1.0}%
{\setlength{\tabcolsep}{5pt}
\begin{NiceTabular}{l |c c c c|c c c c}
\toprule
\multirow{2}{*}{\bf Model} & \multicolumn{4}{c|}{\bf LiveCodeBench-Python} & \multicolumn{4}{c}{\bf LiveCodeBench-C++} \\
& Easy & Medium & Hard & All & Easy & Medium & Hard & All \\
\midrule
DeepSeek-R1                 & 98.5 & 79.8 & 37.4 & 65.6 & 95.5 & 75.3 & 29.3 & 59.9 \\
QwQ-32B                     & 97.0 & 79.8 & 28.5 & 61.3 & 94.0 & 68.5 & 26.0 & 55.9 \\
Qwen3-32B                   & 98.0 & 79.9 & 34.6 & 64.3 & 96.3 & 73.0 & 35.0 & 61.9 \\
\midrule
\multicolumn{8}{c}{\bf Distilled 7B+ Models} \\
\midrule
R1-Distill-Qwen-7B          & 86.6 & 43.8 & 7.0 & 38.0 & 26.9 & 5.6 & 1.6 & 9.0 \\
OpenThinker2-7B             & 80.6 & 16.9 & 1.6 & 25.5 & 43.3 & 2.3 & 0 & 11.1 \\
OlympicCoder-7B             & 82.1 & 49.4 & 12.2 & 40.9 & 85.7 & 46.7 & 10.2 & 40.0 \\
OCR-Qwen-7B-Instruct        & 95.4 & 64.0 & 18.0 & 51.3 & 13.4 & 2.3 & 0.8 & 4.3 \\
\rowcolor{gray!20}{OCR-2-7B} & 97.0 & 71.1 & 20.9 & {\bf 55.2} & 91.4 & 64.5 & 21.2 & {\bf 51.9} \\
\midrule
\multicolumn{8}{c}{\bf Distilled 14B+ Models} \\
\midrule
R1-Distill-Qwen-14B         & 98.5 & 62.9 & 17.1 & 51.3 & 68.7 & 39.3 & 6.5 & 31.9 \\
DeepCoder-14B-Preview       & 97.0 & 65.2 & 19.5 & 52.7 & 61.2 & 39.3 & 8.9 & 39.3 \\
OCR-Qwen-14B-Instruct       & 97.6 & 74.4 & 27.6 & {\bf 59.4} & 47.8 & 16.9 & 0.8 & 17.2 \\
\rowcolor{gray!20}{OCR-2-14B} & 97.9 & 75.4 & 26.8 & {\bf 59.4} & 91.8 & 68.6 & 25.7 & {\bf 55.3} \\
\midrule
\multicolumn{8}{c}{\bf Distilled 32B+ Models} \\
\midrule
R1-Distill-Qwen-32B         & 98.5 & 68.5 & 28.5 & 58.1 & 80.6 & 39.3 & 11.4 & 36.9 \\
OpenThinker2-32B            & 97.0 & 65.2 & 22.8 & 54.1 & 97.0 & 60.7 & 25.2 & 53.8 \\
OlympicCoder-32B            & 98.5 & 71.9 & 24.4 & 57.4 & 91.0 & 62.8 & 21.3 & 51.4 \\
OCR-Qwen-32B-Instruct       & 98.4 & 77.2 & 30.4 & 61.7 & 65.7 & 33.7 & 4.1 & 28.3 \\
\rowcolor{gray!20}{OCR-2-32B} & 97.9 & 77.1 & 31.8 & {\bf 62.1} & 94.7 & 72.2 & 28.0 & {\bf 58.1} \\
\bottomrule
\end{NiceTabular}
}
\vspace{2mm}
\caption{Performance comparison of reasoning models on LiveCodeBench. Highlighted rows show our finetuned models' performances. Bold indicates the highest performance. Python results are averaged across 64 runs, and C++ results across 16 runs.}
\label{tab:main_results}
\vspace{-2mm}
\end{table}

\paragraph{Training and Inference Hyper-Parameters}

By leveraging \ourdataset, we gauged the efficacy of supervised fine-tuning (SFT) through the adaptation of \texttt{Qwen2.5-Instruct} models, spanning parameter counts of 7B, 14B, and 32B. The model training uses AdamW optimizer \citep{kingma2014adam} with a learning rate $5e-5$, a batch size of 256, and a maximum context length of 32,768 tokens. We used a CosineAnnealing learning rate schedule with a 5\% warmup, and the final checkpoint was used for evaluation. To accelerate training, we utilized sequence packing \citep{shen2024nemoaligner}, tensor and context parallelism, and BF16 precision. For generating outputs during inference, we employed temperature-based nucleus sampling \citep{holtzman2020nucleus_sampling} via vLLM \citep{kwon2023vllm}, setting a maximum output length of 30,720 tokens.

\paragraph{Baselines}
The following open-weight models were chosen as baselines in our evaluation: DeepSeek-R1 and R1-Distill-Qwen models \citep{deepseek_r1}, QwQ-32B \citep{qwq32b}, Qwen3-32B \citep{qwen3}, OlympicCoder \citep{openr1_codeforces}, OpenThinker2 \citep{openthoughts}, DeepCoder-14B-Preview \citep{deepcoder2025}, and OCR-Qwen \citep{ahmad2025opencodereasoningadvancingdatadistillation} models.

\paragraph{Benchmarks and Metrics} For our evaluation, we used the same LiveCodeBench \citep{jain2025livecodebench} split that we utilized for our C++ expansion. This benchmark contains 67 easy, 89 medium, and 279 hard coding questions. We report \texttt{pass@1} for code generation, \texttt{pass@1|select@k} under test-time scaling setup. Furthermore, we report the accuracy of the self-critique capability. An LLM's prediction is considered correct if it accurately judges the correctness of all generated solutions ($k$ unless otherwise mentioend) for a given input question.
We also evaluate critique accuracy using CodeContests benchmark \citep{li2022codecontests} and provide details in \autoref{sec:critique_eval}.


\begin{table}[t]
\centering
\def\arraystretch{1.0}%
{\setlength{\tabcolsep}{5pt}
\begin{NiceTabular}{l |c c| c c|c c}
\toprule
\multirow{2}{*}{\bf Model} & \multicolumn{2}{c|}{\bf Pass@1} & \multicolumn{2}{c|}{\bf Pass@10} & \multicolumn{2}{c}{\bf Pass@1|Select@10}  \\
& Python & C++ & Python & C++ & Python & C++ \\
\midrule
DeepSeek-R1                     & 61.3 & 60.1 & 75.3 & 72.0 & 63.4~\bplus{2.1} & 62.7~\bplus{2.6} \\
QwQ-32B                         & 60.2 & 54.1 & 73.5 & 70.3 & 64.2~\bplus{4.0} & 56.3~\bplus{2.2} \\
Qwen3-32B                       & 63.6 & 61.9 & 77.4 & 75.3 & 67.4~\bplus{3.8} & 63.8~\bplus{1.9} \\
\midrule
DeepCoder-14B-Preview           & 53.0 & 29.4 & 65.9 & 54.5 & 57.7~\bplus{4.4} & 35.1~\bplus{\bf 5.7} \\
OpenThinker2-32B                & 58.1 & 51.6 & 74.2 & 68.8 & 62.4~\bplus{4.3} & 53.4~\bplus{1.8} \\
OlympicCoder-32B                & 55.6 & 52.3 & 71.3 & 68.5 & 59.9~\bplus{4.3} & 55.6~\bplus{3.3} \\
\rowcolor{gray!20}{OCR-2-7B}    & 55.2 & 51.8 & 67.7 & 69.9 & 60.2~\bplus{5.0} & 54.1~\bplus{2.3} \\
\rowcolor{gray!20}{OCR-2-14B}   & 58.6 & 56.4 & 72.0 & 69.5 & 60.6~\bplus{2.0} & 58.4~\bplus{2.0} \\
\rowcolor{gray!20}{OCR-2-32B}   & 61.3 & 59.8 & 75.6 & 73.1 & 67.4~\bplus{\bf 6.1} & 60.6~\bplus{0.8}   \\
\bottomrule
\end{NiceTabular}
}
\vspace{2mm}
\caption{Performance comparison of reasoning models under test-time scaling setup. Highlighted rows show our finetuned models' performances. The pass@1 scores are averaged over 10 runs. The performance gains with self-critique are highlighted in blue and bold values indicate the largest gains.}
\label{tab:gen_crit_results}
\vspace{-3mm}
\end{table}

\begin{table}[t]
\centering
\def\arraystretch{1.0}%
{\setlength{\tabcolsep}{5pt}
\begin{NiceTabular}{l |c c c c|c c c c}
\toprule
\multirow{2}{*}{\bf Model} & \multicolumn{4}{c|}{\bf LiveCodeBench-Python} & \multicolumn{4}{c}{\bf LiveCodeBench-C++} \\
& Easy & Medium & Hard & All & Easy & Medium & Hard & All \\
\midrule
DeepSeek-R1                 & 91.0 & 47.2 & 8.9 & 40.9 & 82.1 & 40.4 & 13.0 & 38.4 \\
QwQ-32B                     & 67.2 & 39.3 & 9.8 & 33.0 & 74.6 & 25.8 & 9.8 & 30.5 \\
Qwen3-32B                   & 77.6 & 41.6 & 9.8 & 36.2 & 74.6 & 30.3 & 9.8 & 31.9 \\
\midrule
DeepCoder-14B-Preview      & 10.4 & 1.1 & 0 & 2.9 & 4.5 & 1.1 & 0 & 1.4 \\
OpenThinker2-32B           & 11.9 & 2.2 & 0.8 & 3.9 & 35.8 & 1.1 & 0.8 & 9.3 \\
OlympicCoder-32B           & 56.7 & 39.3 & 4.9 & 28.3 & 59.7 & 27.0 & 0.8 & 23.3 \\
\rowcolor{gray!20}{OCR-2-7B} & 80.6 & 47.2 & 14.6 & 40.9 & 70.1 & 24.7 & 8.9 & 28.7 \\
\rowcolor{gray!20}{OCR-2-14B} & 88.1 & 43.8 & 16.3 & 42.3 & 82.1 & 36.0 & 8.9 & 35.1 \\
\rowcolor{gray!20}{OCR-2-32B} & 92.5 & 53.9 & 17.1 & {\bf 47.0} & 80.6 & 44.9 & 15.5 & {\bf 40.5} \\
\bottomrule
\end{NiceTabular}
}
\vspace{2mm}
\caption{Accuracy comparison of reasoning models on self-critique. Highlighted rows show our finetuned models' performances. Bold indicates the best performance. }
\label{tab:critique_results}
\vspace{-4mm}
\end{table}

\subsection{Main Results}
Tables~\ref{tab:main_results} and \ref{tab:gen_crit_results} summarize the performance of our distilled models against various competing baselines. We consistently observed the following three trends.

\paragraph{Scaling Yields Large Performance Boosts for Smaller Models}
A comparison between our OCR-2 models and the previous OCR-Qwen models in Table~\ref{tab:main_results} demonstrates that scaling the quantity of synthetic solutions particularly benefits smaller models. Increasing the fine-tuning samples from 737K to 2.5M yields a more substantial relative performance improvement for smaller models (e.g., 7B parameters) compared to their larger counterparts. This trend is also evident in C++ performance, especially when compared to models like OlympicCoder, which were trained on C++ data. Although the 32B parameter model also improves, the gains suggest it might be approaching a performance ceiling achievable through mere scaling of synthetic data quantity for existing problem types. 

\paragraph{LLMs Show Similar Capabilities in Python and C++}
Notably, \ourdataset includes approximately 1.4M Python samples and 1.17M C++ samples as seen in Table~\ref{tab:data_stat}. Consequently, our models maintain or improve their Python capabilities while demonstrating dramatically enhanced C++ solution quality. Despite being trained on both languages, our models exhibit significantly superior performance on LiveCodeBench-C++ compared to other models of similar size. In particular, the 32B parameter model we train exceeds QwQ-32B and other open-weight reasoning models in both Python and C++. Overall, joint training on substantial, comparably-sized Python and C++ solution sets results in strong performance in both languages, often outperforming models trained on single-language data. We anticipate this positive transfer learning behavior could extend to other programming languages, a direction we leave for future work.

\paragraph{Test-time Scaling with Self-Critique Yields Significant Improvements}
Table~\ref{tab:gen_crit_results} highlights the benefits of applying self-critique at test-time, a capability developed by fine-tuning models on data that includes self-critique labels. For instance, when considering our flagship OCR-2-32B model, using its self-critique ability to select the best solution advances the Pass@1 score by approximately 6 percentage points. This enhancement significantly narrows the performance gap between Pass@1 and Pass@10 for all our model sizes, in both Python and C++, and results in our models outperforming other similarly sized competitors. Therefore, training with self-critique data not only improves baseline Pass@1 scores but also demonstrates that self-critique is an effective test-time strategy for selecting a higher accuracy solution from multiple parallel generations.

\section{Ablation and Analyses}
\label{sec:ablation_and_analysis}

\subsection{Test-time Scaling: Opportunities for Enhanced Critique}
\begin{figure}[t]
\centering
\includegraphics[width=\textwidth]{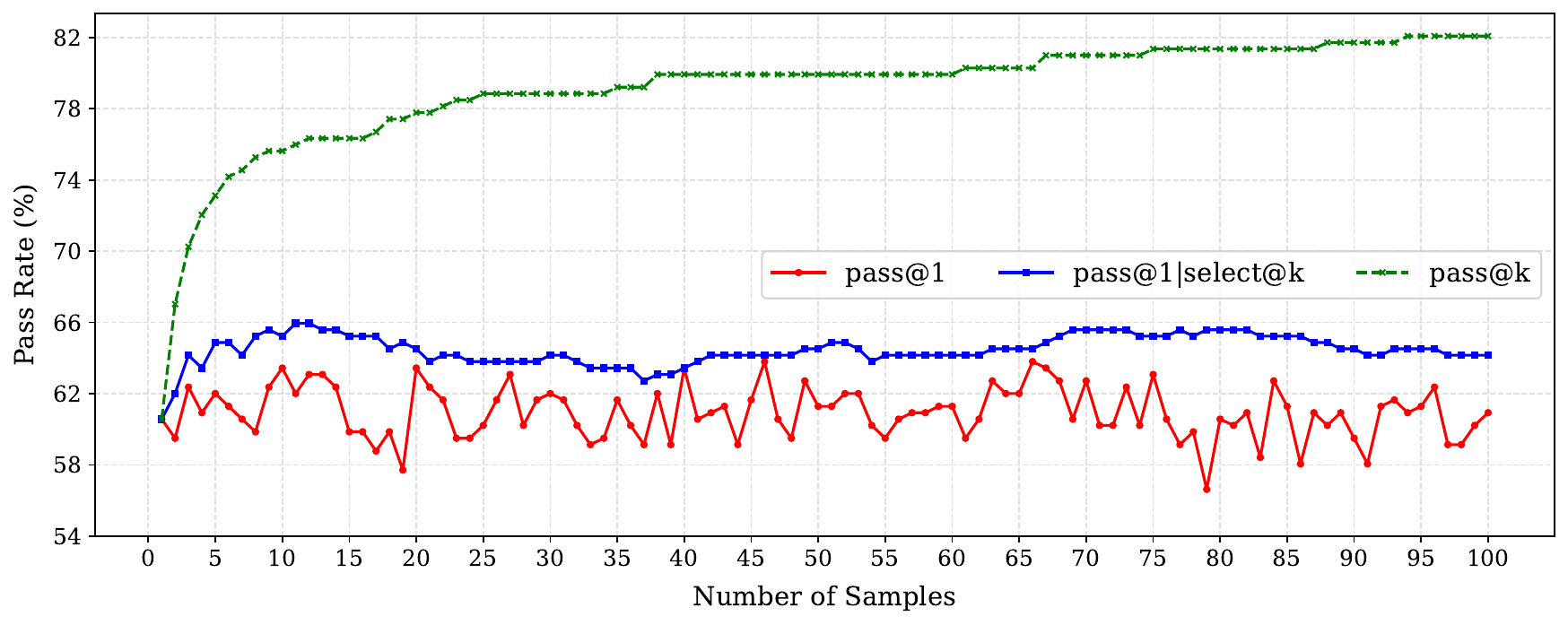}
\caption{Performance gap between \texttt{pass@1}, \texttt{pass@1|select@k}, and \texttt{pass@k} under test-time scaling - large number of samples drawn from \textbf{OCR-2-32B}.}
\label{fig:pass_rate_comparison}
\vspace{-2mm}
\end{figure}

Although self-critique based selection under parallel test-time scaling has proven effective in boosting LLMs' performance with a limited number of samples, an important question left to address is, what is the gap between a model's \texttt{pass@1}, \texttt{pass@1|select@k}, and \texttt{pass@k} as k grows? In this ablation, we aim to answer this question using a far larger value of k (up to 100) and show the gap that exists between them. It can be seen in \autoref{fig:pass_rate_comparison} that \textit{OCR-2-32B} quickly attains a high pass@k score. For comparison, we plot the pass@1 score for each individual sample, computed independently from the rest of the samples, as well as the pass@1|select@k score, as $k$ increases to include all past solutions. 

Two important observations can be drawn from this figure. First, while pass@1|select@k is consistently higher than pass@1, this increase does not go beyond a certain limit. This observation is consistent with the fact that out of the 279 problems in LiveCodeBench split we evaluate on, nearly 212 problems fall under the Medium/Hard category where self-critique accuracy is insufficient to correctly determine the correctness of the all generated samples, as can be seen in \autoref{tab:critique_results}.
On evaluating the accuracy of the critique labels using various models on a single solution for each problem, we find that the accuracy of these models for medium difficulty problems tends to be 47\% and for hard problems is less than 14\% at best when critiquing 10 solutions. This inaccurate final determination of correctness label limits the scope of improvement on the overall benchmark, as the vast majority (75.9\%) of samples in LiveCodeBench are inaccurately critiqued. Secondly, the simple heuristic described in \autoref{sec:method}, which is to select the solution with shortest critique reasoning trace, may be ineffective in incorporating information from additional solutions and prevents substantial improvements. We leave the exploration of more sophisticated heuristics and methods to enhance the accuracy of self-critique-based selection for future research.


\subsection{Impact of Temperature on Self Critique}


We tested how the critic LLMs responded to different decoding temperatures by re-evaluating them at temperature $[0.2, 0.4, 0.6, 0.7]$. The pass@1|select@k and the accuracy of self-critique showed minimal variation (less than 3\%) across these temperatures, and t-tests indicated no statistically significant differences. These findings suggest that within the range below 0.7, the decoding temperature does not substantially affect the critic's performance. This implies that the model's judgments are mainly based on its internal knowledge rather than the randomness of the sampling process. Therefore, we used a temperature of 0.6 throughout this work.

\subsection{Transfer Learning: Python $\leftrightarrow$ C++}

\begin{wraptable}{r}{70mm}
\centering
\vspace{-3mm}
\def\arraystretch{1.0}%
\resizebox{70mm}{!}{
\begin{tabular}{ l | c c | c c}
\toprule
\multirow{2}{*}{\textbf{Model}} & \multicolumn{2}{c|}{\textbf{Dataset Size}} & \multicolumn{2}{c}{\textbf{LiveCodeBench}} \\
& \textbf{Python} & \textbf{C++} & \textbf{Python} & \textbf{C++} \\ 
\midrule
\multirow{3}{*}{OCR-2-7B}      & 1.4M  & 0      & 54.9 & \textcolor{red}{\bf 1.7} \\ 
                                & 0     & 1.1M  & 36.0 & 45.3 \\ 
                                & 1.4M  & 1.1M  & 55.2 & 51.9 \\
\arrayrulecolor{gray!70} \midrule \arrayrulecolor{black}                           
\multirow{3}{*}{OCR-2-14B}      & 1.4M  & 0     & 56.3 & \textcolor{red}{\bf 9.8} \\ 
                                & 0     & 1.1M  & 54.0 & 53.6 \\ 
                                & 1.4M  & 1.1M  & 59.4 & 55.3 \\ 
\arrayrulecolor{gray!70} \midrule \arrayrulecolor{black}
\multirow{3}{*}{OCR-2-32B}      & 1.4M  & 0     & 60.5 & \textcolor{red}{\bf 16.0} \\ 
                                & 0     & 1.1M  & 56.6 & 56.4 \\ 
                                & 1.4M  & 1.1M  & 62.1 & 59.4 \\ 
\bottomrule
\end{tabular}
}
\caption{Pass@1 scores of OCR-2 models trained individually on Python and C++ vs. jointly using \ourdataset.}
\label{tab:lang_transfer}
\end{wraptable}

\ourdataset features solutions in both Python and C++, allowing study to investigate how well models perform when trained on a single language and tested on the other. The results of these experiments are detailed in  \autoref{tab:lang_transfer}. First, our findings suggest that cross-language transfer does occur, and combining Python and C++ data during training enhances overall performance on both languages. Second, we observe an asymmetry: models trained solely on C++ achieve noticeable scores on Python, whereas models trained only on Python, while performing well on Python, experience a significant drop in accuracy on C++. This difference isn't simply due to dataset size, and further research is needed to understand the underlying causes of this performance degradation.

\subsection{Impact of Scaling Up Data on Code Generation}
\label{sec:scaling_data}

\begin{figure*}[t]
\centering
\includegraphics[width=0.98\textwidth]{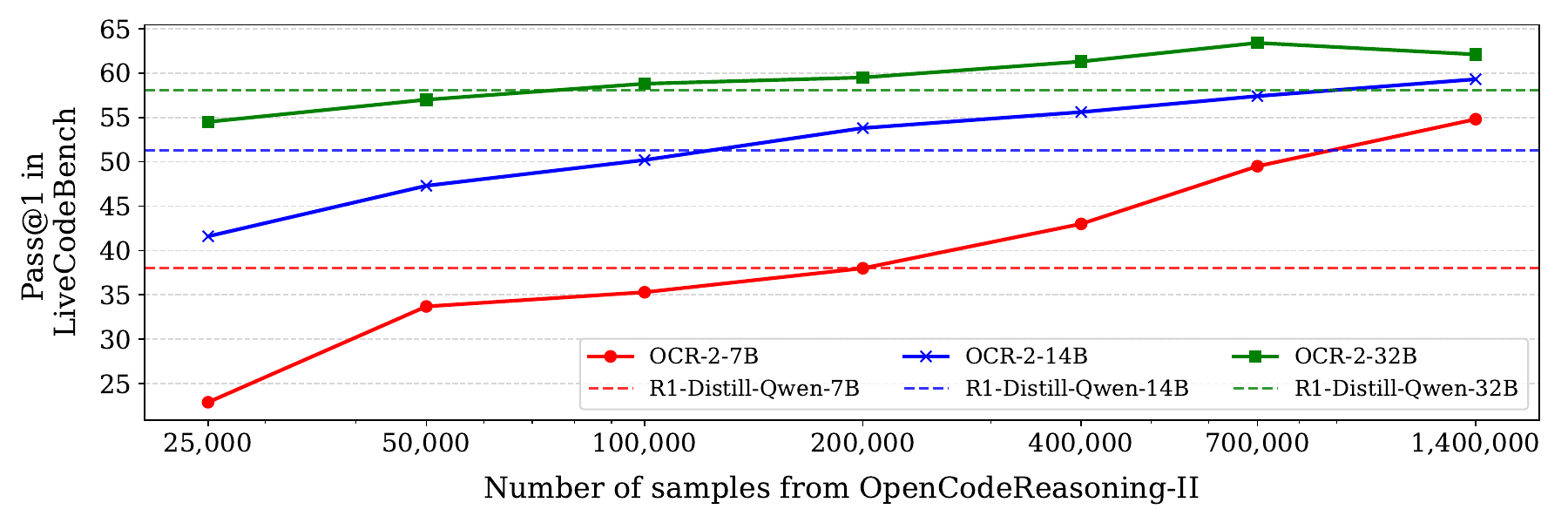}
\vspace{-2mm}
\caption{ Impact of scaling up data from 25k to 1.4M samples in \ourdataset.}
\label{fig:oci_overview}
\vspace{-2mm}
\end{figure*}

We analyze the data scaling study of \textit{OpenCodeReasoning} \citep{ahmad2025opencodereasoningadvancingdatadistillation}, and substantially increase the dataset size in order to determine whether data scaling shows limits for a given model size. As such, we redo the scaling study from 25K samples all the way to 1.4M samples in Python using \ourdataset and plot the trajectory of scores on LiveCodeBench in \autoref{fig:oci_overview}. While the 7B model shows substantial improvements in score with data scaling, such improvement is not observed in the 14B and 32B models. It remains to be seen if access to more novel and complex instructions may further improve scores, or if the number of solutions to the existing problems must be scaled by orders of magnitude to improve scores further measurably.

\section{Related Works}
\label{sec:related_works}

Our work builds upon a foundation of research in synthetic data generation for code, the growing reasoning capabilities of LLMs, and model-based critique, particularly within the coding context. The demonstrated effectiveness of LLMs in coding has spurred the creation of numerous impactful synthetic datasets for instruction tuning \citep{wang-etal-2023-self-instruct, wei2023magicoder, xu2024wizardlm, wu2024inversecoder, luo2024wizardcoder, wei2024selfcodealign, majumdar2024genetic, ahmad2025opencodeinstructlargescaleinstructiontuning}. Inspired by the positive impact of extended inference-time computation on code quality, synthetic data generation has also been adapted for training LLMs with reasoning abilities for code. These efforts have shown substantial gains from fine-tuning on as few as 17k reasoning-style CoT examples \citep{li2025llmseasilylearnreason} and scaling up to 447k \citep{bespoke_stratos, penedo2025codeforces, openthoughts, li2025llmseasilylearnreason, xu2025kodcodediversechallengingverifiable}. Recent work by \citet{ahmad2025opencodereasoningadvancingdatadistillation} further scaled this data distillation strategy to 737k samples, achieving superior SFT performance in code and reasoning foundation models \citep{bercovich2025llamanemotronefficientreasoningmodels}. In this paper, we aim to push the boundaries of synthetic data scaling for coding and, crucially, explore critique fine-tuning for reasoning data distillation.

Training reward models to critique solutions is a well-explored research area across various domains. Generalist reward models, trained on preference pairs, have shown strong performance in alignment and reasoning \citep{nvidia2024nemotron4340btechnicalreport, liu2024skywork, wang2024helpsteer2, wang2024helpsteer2preferencecomplementingratingspreferences}. Specialized reward models have also been developed for tasks like math or coding \citep{wang2024mathshepherdverifyreinforcellms, zeng2025acecoderacingcoderrl, liu2025acemathadvancingfrontiermath, zhang2025lessonsdevelopingprocessreward, yang2024qwen25mathtechnicalreportmathematical}. Notably, some of these works have released their training datasets, providing valuable resources for the research community.

LLM-based solution critiques, when combined with reasoning, significantly enhance model capabilities in both mathematical and coding tasks. Increased test-time computation consistently improves verifiers like reward models and test case generators \citep{mahan2024generativerewardmodels, ficek2025scoringverifiersevaluatingsynthetic, liu2025inferencetimescalinggeneralistreward, chen2025rmr1rewardmodelingreasoning, moshkov2025aimo2winningsolutionbuilding}, as demonstrated by \citet{zhang2025generativeverifiersrewardmodeling}'s improved math reward model and released CoT data. Moreover, critique fine-tuning positively impacts question-answering \citep{sun-etal-2024-critique, yu-etal-2025-self}. This synergy, further explored in recent work combining LLM critiques with reasoning-driven test-time scaling, offers a promising route to advance coding abilities \citep{wang2025critiquefinetuninglearningcritique, zhou2025refinecoderiterativeimprovinglarge}.

\section{Conclusion}
\label{sec:conclusion}

This research addresses the critical need for high-quality large-scale data to propel advancements in reasoning-based LLMs for test-time scaling. We introduced \ourdataset, a significantly larger and richer dataset of question-solution-critique triples that has enabled us to train powerful distilled models. Our two-stage fine-tuning approach yielded \texttt{Qwen2.5-Instruct} models that demonstrate state-of-the-art or comparable code generation capabilities among open-weight distilled models. More importantly, the synergistic combination of our code generation and critique models led to tangible improvements in competitive coding benchmarks. Additionally, our C++ extension of LiveCodeBench broadens the evaluation landscape for LLMs in code.

\bibliography{bib/anthology,bib/custom}
\bibliographystyle{elsarticle-harv}

\appendix
\appendix
\clearpage
{
\centering
\Large\bf Technical Appendices and Supplementary Material \\ [20pt]
}

\section{Final Output Selection using Critique}
\label{sec:critique_selection_ablation}

\begin{figure*}[h]
\centering
\includegraphics[width=\textwidth]{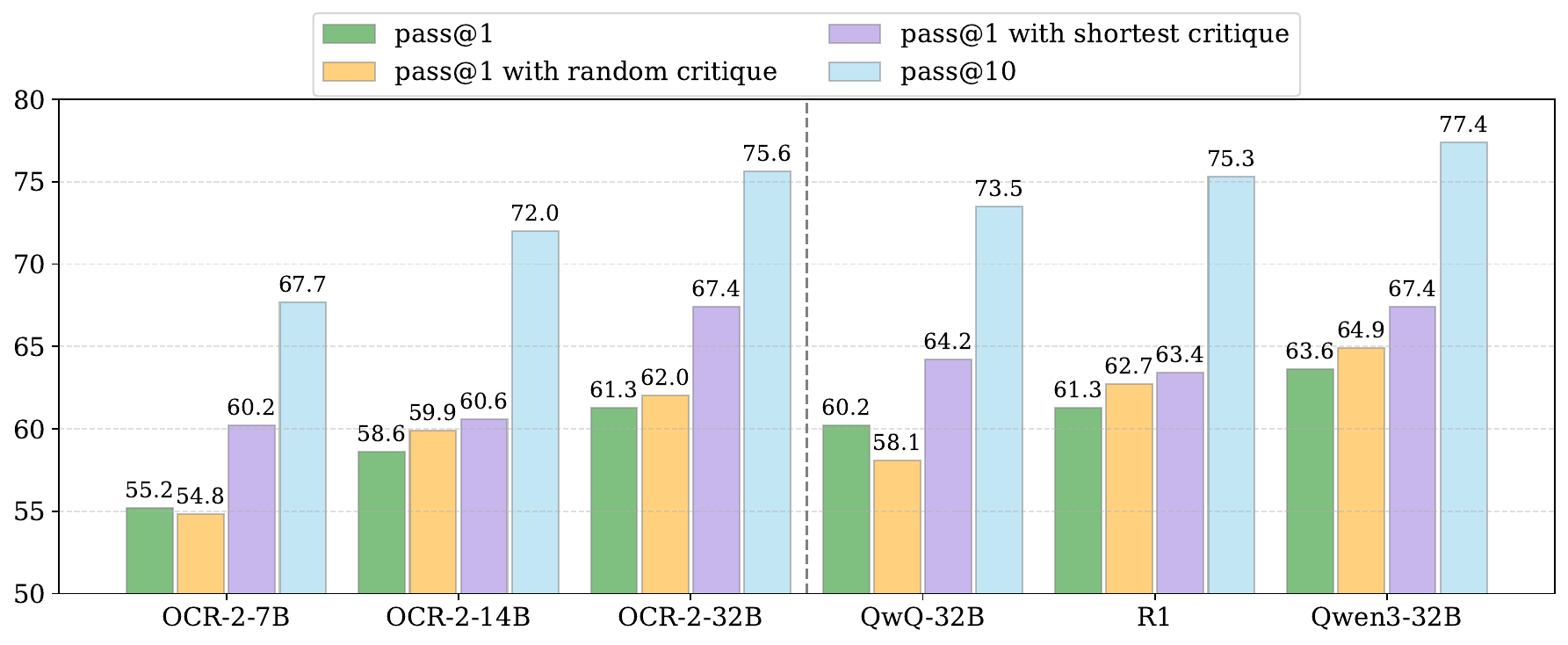}
\caption{Differences in pass@1 scores between randomly selecting the final output vs. choosing right solution with shortest critique thinking, using OCR-2-32B on LiveCodeBench-Python.}
\label{fig:critique_selection_comparison}
\end{figure*}

To calculate pass@1, we select the shortest reasoning trace of the critique. A straightforward baseline for final solution selection is uniform random selection. Figure \ref{fig:critique_selection_comparison} contrasts these two selection methods, revealing that our heuristic for choosing the shortest trace yields considerably better scores than randomly picking from the positive critique samples. We hypothesize that the critique model's comparatively lower accuracy in identifying correct solutions for \texttt{medium} and \texttt{hard} problems explains the substantially weaker performance of the random selection baseline.

\section{Evaluating Critique LLM Accuracy in Judging Code Solutions}
\label{sec:critique_eval}

This study initially assessed LLMs' self-critique ability by having them evaluate their own generated solutions to select the best output in a parallel scaling setup. Recognizing that this might not reflect their true critique capabilities due to varying generation accuracy, we further evaluated them as external critics. To do this, we used the CodeContests benchmark \citep{li2022codecontests}, randomly selecting one correct and one incorrect human-written solution for each of its 165 test questions. We limited code solutions to a maximum of 4096 tokens (based on the \texttt{Qwen2.5} tokenizer), resulting in 238 Python and 329 C++ samples. The critique performance results when calculated on single positive-negative sample per problem are shown in \autoref{tab:cc_critique_eval}. Notably, while OCR-2-32B excelled in critiquing Python code, OCR-2-14B surprisingly achieved the best performance for C++.


\begin{table}[ht]
\centering
\def\arraystretch{1.0}%
\begin{tabular}{l |c c |c c}
\toprule
Model & \# Language & \# Accuracy & \# Language & \# Accuracy \\
\midrule
DeepCoder-14B-Preview   & Python & 13.4 & C++ & 28.6 \\
OpenThinker2-32B        & Python & 43.7 & C++ & 45.9 \\
OlympicCoder-32B        & Python & 46.6 & C++ & 49.5 \\
QwQ-32B                 & Python & 60.1 & C++ & 56.8 \\
Qwen3-32B               & Python & 63.9 & C++ & 65.0 \\
\midrule
OCR-2-7B                & Python & 60.1 & C++ & 54.7 \\
OCR-2-14B               & Python & 63.9 & C++ & {\bf 65.3} \\
OCR-2-32B               & Python & {\bf 66.0} & C++ & 65.0 \\
\bottomrule
\end{tabular}
\vspace{2mm}
\caption{Critique accuracy of reasoning-enabled LLMs on human-written solutions provided in the test split of Code-Contests benchmark \citep{li2022codecontests}.}
\label{tab:cc_critique_eval}
\end{table}

\newpage












\begin{figure*}[ht!]
    \centering
    \begin{tcolorbox}[title={Prompt for solution generation in Python}, colback=red!0, left=2pt,right=2pt,top=0pt,bottom=0pt]
        { 
        \vspace{0.1cm}
        system: ""

        \vspace{0.1cm}
        user: |-

        \quad You are a helpful and harmless assistant. You should think step-by-step before responding to the instruction below. \\

        \quad Please use python programming language only. \\

        \vspace{0.1cm}
        \quad You must use \threebackticks python for just the final solution code block with the following format:

        \quad\threebackticks python

        \quad\# Your code here

        \quad\threebackticks

        \quad\{input\} \\

        }
    \end{tcolorbox}

    \vspace{0.25cm} 

    \begin{tcolorbox}[title={Prompt for solution generation in C++}, colback=red!0, left=2pt,right=2pt,top=0pt,bottom=0pt]

    { 
    \vspace{0.1cm}
    system: ""
    
    \vspace{0.1cm}
    user: |-
    
    \quad You are a helpful and harmless assistant. You should think step-by-step before responding to the instruction below. \\
    
    \quad Please use c++ programming language only. \\
    
    \vspace{0.1cm}
    \quad You must use \threebackticks cpp for just the final solution code block with the following format:
    
    \quad\threebackticks cpp
    
    \quad\# Your code here
    
    \quad\threebackticks
    
    \quad\{input\} \\
    
    }
    \end{tcolorbox}
    \caption{Prompt template used for solution generation using R1 for \ourdataset.}
    \label{fig:combined_data_gen}
\end{figure*}

\begin{figure*}[ht!]
\centering

\begin{tcolorbox}[title={Prompt for critique generation using QwQ-32B}, colback=red!0, left=2pt,right=2pt,top=0pt,bottom=0pt]

{ 
\vspace{0.1cm}
system: ""

\vspace{0.1cm}
user: |-

\quad You are a helpful and harmless assistant. You should think step-by-step before responding to the instruction below. \\

\quad You have solved a programming problem. Now, you will critique your solution and conclude with <judgment>right/wrong</judgment>. \\

\vspace{0.1cm}
\quad\#\# Question

\quad\{question\}  \\

\quad\#\# Solution

\quad\{solution\} \\

}
\end{tcolorbox}
\caption{Prompt template used for critique data generation for \ourdataset.}
\label{fig:critique_data_gen}
\end{figure*}

\end{document}